\documentclass[]{interact}

\usepackage{epstopdf}
\usepackage{subfigure}
\usepackage{multirow}
\usepackage{algorithm}
\usepackage{algorithmic}

\usepackage[numbers,sort&compress]{natbib}
\bibpunct[, ]{[}{]}{,}{n}{,}{,}
\renewcommand\bibfont{\fontsize{10}{12}\selectfont}

\theoremstyle{plain}

\theoremstyle{definition}

\theoremstyle{remark}

\begin{document}
	
	\articletype{ARTICLE}
	
	\title{Differential Privacy for Sparse Classification Learning}
	
	\author{
		\name{Puyu Wang \textsuperscript{a} , Hai Zhang \textsuperscript{a,b}  \thanks{CONTACT Hai Zhang. Email: zhanghai@nwu.edu.cn} }
		\affil{\textsuperscript{a}School of Mathematics, Northwest University,
			Xi'an; \textsuperscript{b}Faculty of Information Technology  \& State Key Laboratory of Quality Research in Chinese Medicines, Macau University of Science and Technology, Macau}
	}
	\thanks{This work was partially supported by National Natural Sciences China-Guangdong Joint Fund under grant numbers U1811461, National Natural Science Foundation of China under grant numbers 11571011.}

	\maketitle
	
	\begin{abstract}
		
In this paper, 
we present a differential privacy version of convex and nonconvex sparse classification approach. 
Based on alternating direction method of multiplier (ADMM) algorithm, 
we transform the solving of sparse problem into the multistep iteration process. 
Then we add exponential noise to stable steps to achieve privacy protection. 
By the property of the post-processing holding of differential privacy, 
the proposed approach satisfies the $\epsilon-$differential privacy even when the original problem is unstable. 
Furthermore, we present the  theoretical privacy bound of the differential privacy classification algorithm. Specifically, 
the privacy bound of our algorithm is controlled by the algorithm iteration number, the privacy parameter, the parameter of loss function, ADMM pre-selected parameter, and the data size. 
Finally we apply our framework to logistic regression with  $L_1$ regularizer and logistic regression with $L_{1/2}$ regularizer. 
Numerical studies  demonstrate that our method is both effective and efficient which performs well in sensitive data analysis.

	\end{abstract}
	
	\begin{keywords}
		
	Differential privacy; Classification; ADMM; Unstable; Sparse
	\end{keywords}

	\section{Introduction}

	With the development of the internet of things, 
	a large amount of data has been collected which contains the individuals sensitive information, 
	such as medical purchase records, hospital electronic medical records,
	web-site search information, home address and contact information.  
	Inappropriate use of sensitive information will cause a serious threat to personal privacy, which will bring the risk of personal privacy leakage. 
	For example, 
	an attacker can infer personal information or predict individual behavior by capturing personal sensitive behavior data easily.    
	In order to avoid the leakage of privacy, scholars have proposed many different privacy-preserving approaches.
	Sweeney et al.\cite{Sweeney1, Sweeney2} proposed K-anonymous method,
	Machanavajjhala et al. \cite{Machanavajjhala} proposed $l-$diversity principle,
	Li et al.\cite{Li} proposed T-closeness and so on. 
	These technologies have been studied extensively and applied to privacy protecting. 
	However, the reliability of these privacy protection models is related to the background knowledge of the attacker. 
	When the attacker has enough background knowledge, these methods may fail\cite{Ganta}\cite{Wong}.
	Dwork \cite{Dwork} proposed differential privacy in 2006.
	Unlike traditional privacy-preserving methods,
	differential privacy defines a strict attack model which is independent of background knowledge,
	and gives a quantitative representation of the degree of privacy leakage,
	which can protect individual privacy information effectively. In the seminal paper \cite{Dwork}, Dwork gave the definition of differential privacy, which can be stated formally as follows.
	
	\
	
	\noindent \textbf{Definition\ }
	Given a random algorithm $M$,
	$Range(\cdot)$ denote a collection of all possible outputs of algorithm.
	For any two data sets $X$ and $X^{'}$ that differing on at most one data point,
	and for any $S\subseteq Range(M)$, if
	$$ \frac{P(M(X))\in S )} {P( M(X^{'})\in S)} \leq \exp(\epsilon), $$
	then random algorithm $M$ satisfies $\epsilon-$differential privacy.

	\
	
	As a privacy-preserving technology with rigorous mathematical theory,
	differential privacy is suitable for privacy preserving in the era of big data, 
	and has a broad application \cite{Dwork1,Liu,Sui,Brinkrolf}. 
	In the framework of differential privacy, the degree of privacy information leakage is controlled by the parameter $\epsilon$. 
	As the $\epsilon$ increases, the more privacy is revealed, which means that the probability  of individual indentification increases. 
	The advantages of differnetial privacy mainly include: (1) Differential privacy has good privacy protection ability even if attacker has background knowledge, because the attack model of differential privacy is independent of the background knowledge.
	(2) Differential privacy has the post-processing property, specifically,  
	if data published satisfies a given amount $\epsilon-$differential privacy, the privacy leakage will not increase with data analyzing or re-releasing. 
	The above advantages make differential privacy have good scalability.
	Laplace mechanism\cite{Dwork3} and exponential mechanism\cite{Mcsherry}
	are two common differential privacy mechanisms.
	Although differential privacy has a wide range of applications, 
	most of applications focus on data publishing \cite{Bolot, Karwa}
	and query processing \cite{Cormode, Li1}.

	In this paper, we focus on the regularized empirical risk minimization 
	(ERM) problems for binary classification, which is defined as follows
	$$\underset{w}{\min}\ \frac{1}{n}\sum_{i=1}^nO(y_{i}w^Tx_{i})+P_{\lambda}(w),\eqno(1)$$
	where $O(\cdot)$ is loss function, $x_i\in \textbf{R}^p$, $y_i\in \{-1,1\}$, $w \in \textbf{R}^p$ is parameter coefficient vector, $P_{\lambda}(w)$ is penalty function, and $\lambda > 0$ is the tunning parameter which controls the complexity of the model. 
	Specifically, we focus on the $L_q$ regularizer, which have the form $P_{\lambda}(w)=\|w\|_q^q$, where $0<q\leq 1$. 
	When $q=1$, it is $L_1$ Lasso type\cite{Tibshirani}, and when
	$q=1/2$, it is $L_{1/2}$ regularizer\cite{Xu}.
	Our goal is to train a sparse classifier $w^{*}$ over the dataset while protect the individual privacy. 
	
	Differential privacy requires the output of the random algorithm being stable under small perturbations of the input data\cite{Dwork4}.  
	However, the most sparse algorithms are unstable.
	Xu\cite{Xu2} studied the relationship between sparsity and stability. Their results showed that a sparse algorithm can not be stable and vice versa. Thus, the unstable property of sparse algorithm  makes it hard to design a differential privacy sparse algorithm. 
	Under some assumptions, 
	there are some works on the convex sparse regularization approaches. 
	For example, Kifer et al.\cite{Kifer} gave the first results on private sparse regression in high dimension. They designed a computationally efficient algorithm, implicitly based on subsampling stability, for support recovery using the LASSO estimator. Smith et al.\cite{Smith} extended and improved on the results of \cite{Kifer}, based on the sufficient conditions for the LASSO estimator to be robust to small changes in the data set. They proposed the privacy algorithms for sparse linear regression. \cite{Talwar}  designed a nearly optimal differentially private version of Lasso. As compared to the previous work, they assumed that the input data has bounded $l_{\infty}$ norm.
	In this paper, we focus on differential privacy sparse classification algorithms.
	Chaudhuri et al.\cite{Chaudhuri} proposed the regularized classification method with differential privacy. Their results hold for loss functions and regularizers satisfying certain differentiability and convexity conditions, specifically, when regularizer is $l_2$ norm.   
	Zhang and zhu\cite{Zhang2} proposed differential privacy regularized classification methods for distributed stored data. Their mechanisms lead to algorithms that can provide privacy guarantees under mild conditions. Zhang et al.\cite{Zhang3} extended and improved the results of \cite{Zhang2}, and based on modified ADMM proposed a perturbation method  to improve privacy without compromising accuracy. Wang and Zhang\cite{Wang} proposed distributed differential privacy logistic regression based on three step ADMM by noising the output of each iteration, which can protect the local privacy. However, the above methods both require the regularizer being stronly-convex and differentiable.

	In this paper, 
	we design a differential privacy version of convex and nonconvex sparse classification approach with mild conditions on the regularizers. 
	To solve the optimization classification problem, we transform the solving process into three-step sub-problems that can be easily solved by iteration process. And only at the second step, the algorithm access the raw data. Interestingly, the algorithm in this step is stable which fits to the tramework of differential privacy.  
	So we design a privacy preserving algorithm. 
	And by the property of the post-processing holding of differential privacy, the proposed approach satisfies the $\epsilon$-differential privacy even when the original problem is unstable.
	Then, we present the theoretical privacy bound of the classification algorithm. 
	At last, we apply our framework to logistic regression with $L_1$ regularizer and logistic regression with $L_{1/2}$ regularizer. 
	Numerical studies  demonstrate that our method is both effective and  efficient which performs well in sensitive data analysis.

\section{Differential privacy Sparse Classification Framework}

In this section, we present the differential privacy sparse classification framework based on ADMM algorithm, then we show that under mild assumptions for loss function and penalty function, our method satisfies $\epsilon-$differential privacy.

\subsection{ADMM algorithm}

ADMM was proposed in the early 1970s, and has since been studied extensively. In recent years, it has been used in many areas such as computer vision, signal processing and networking\cite{Boyd}. 
Now, we use ADMM to  transform the solving of sparse problem into the multistep iteration process. First, consider adding the auxiliary variable $Z\in \textbf{R}^p$, then model (1) can be rewritten as

$$\underset{w,z}{\min}\ \frac{1}{n}\sum_{i=1}^nO(y_{i}w^Tx_{i})+P_{\lambda}(Z),\eqno(2) $$
$$s.t.\ w=Z.$$
It's obvious that model (1) and model (2) are equivalent. 
Then we have the quadratically augmented Lagrangian function
$$
L(Z,w,V)=\frac{1}{n}\sum_{i=1}^nO(y_{i}w^Tx_{i})+
P_{\lambda}(Z) +\frac{c}{2}(\|Z-w+\frac{V}{c}\|_2^2 
-\|\frac{V}{c}\|_2^2), 
$$
where $V$ is Lagrange multipliers, and $c>0$ is a preselected penalty coefficient. 

ADMM algorithm entails three steps per iteration 

\

\noindent Step 1: $Z$ updates,
\begin{align}
Z(k+1)&=\underset{Z}{\arg\min}\ L(Z,w(k),V(k)) \nonumber\\
&=\underset{Z}{\arg\min}\ P_{\lambda}(Z)+\frac{c}{2}(\|Z-w(k)+\frac{V(k)}{c}\|_2^2).  \tag{3}
\end{align}

\noindent Step 2: $w$ updates,
\begin{align}
w(k+1)&=\underset{w}{\arg\min}\ L(Z(k+1),w,V(k)) \nonumber\\
&=\underset{w}{\arg\min}\  \frac{1}{n}\sum_{i=1}^nO(y_{i}w^Tx_{i})+\frac{c}{2}(\|Z(k+1)-w +\frac{V(k)}{c}\|_2^2)  . \tag{4}
\end{align}

\noindent Step 3:\ Lagrange multiplier $V$ updates,
$$V(k+1)= V(k)+c(Z(k+1)-w(k+1)).\eqno(5)$$

\noindent For detials,  
in iteration $k+1$, 
we update $Z(k+1)$ via minimum $L(Z,w(k),V(k))$ with respect to $Z$, 
update $w(k+1)$ via minimum $L(Z(k+1),w,V(k))$ with respect to $w$, 
and update Lagrange multiplier via $V(k+1)=V(k)+c(Z(k+1)-w(k+1))$, 
until the algorithm converges.  

To establish the convergence of ADMM, we have the following assumptions on the data, loss function and regularizer.

\

\noindent \textbf{Assumption 1\ } Suppose that $\|x_{i}\|_2 \leq 1$ for all $i=1,...,n.$

\

\noindent \textbf{Assumption 2\ } Loss function $O(\cdot)$ is strongly convex and Lipschitz differentiable, with $|O^{'}|\leq c_1$, and $0\leq O^{''}\leq c_2$, where $c_1$ and $c_{2}$ are constants. 

\

\noindent \textbf{Assumption 3\ } Regularizer has the form $P_{\lambda}(Z)=\sum_{i=1}^{p}p_{\lambda}(Z_i)$, where $p_{\lambda}(\cdot)$ is restricted prox-regular, that is, if for any $M>0$ and bounded set $T \subseteq dom(f(\cdot))$, there exist $\gamma>0$ such that
$ 
\forall z_2 \in T \backslash S_M,\ d \in \partial f(z_1), \ \|d\|\leq M, $
$$f(z_{1})+\frac{\gamma}{2}\|z_2-z_1\|_2^2 \ge f(z_{2})+\left \langle d, z_1-z_2 \right \rangle, $$
where $S_M:=\{Z_i \in dom(f(\cdot)): \|d\|>M \ for \ all \  d \in \partial f(\cdot)  \}.$

\

\noindent \textbf{Remark 1\ }
In fact, the fitting loss functions include logistic function, and other smooth classification functions, and the regularizers can be some sparsity-inducing functions such as $L_1$ regularizer, and $l_q$ quasi-norms for $0 < q < 1$.  

\

Now, we present the convergence of the above ADMM algorithm.

\

\noindent \textbf{Theorem 1\ } If preselected parameter $c$ satisfies suitable conditions, the sequence generated by the ADMM algorithm (3),\ (4),(5) converges to the stationary point of $L(Z,w,V)$. 

\

\noindent \textbf{Remark 2\ }
Theorem 1 can be considered as a special case of \cite{Wang2}. 
This paper analyzed the convergence of ADMM for minimizing possible nonconvex objective problem. 
Under Assumption 1-3, our sparse classification problem satisfies the convergence conditions A1-A5 in \cite{Wang2}, so Theorem 1 holds.

\

Although ADMM algorithm dose not provide faster convergence compared to other gradient-descent algorithms, it has other advantages. ADMM transform  solving of the original optmization problem into solving two simple  sub-optimization problems. In Step 1, we only need to solve an univariate sparse regression which can be solved by  the existing sparse algorithm. 
In Step 2, we solve a $L_2$ regularization problem, which is easy to be handled. More interestingly, the properties of ADMM are useful for obtaining a private version of the sparse calssification model. 
Notice that in Step 1 of ADMM, the algorithm will never call the training data set, thus it will not cause any leakage of individual privacy,  
although in this step, the algorithm is unstable due to the sparse regularizer. In step 2, the algorithm will call the training data, but now, the objective function is convex and stable which is suitable for the framework of differential privacy. Step 3 is also data-independent, which will not call the training data in each iteration.

Based on the discussion of this subsection, 
it's no longer difficult to study the differential privacy for sparse algorithm.
In the next subsection, 
we propose the differential privacy sparse classification method by noising the data-related sub-optimization problem (4). 
By the post-processing property of differential privacy, 
the update of $Z$ and $V$ in each iteration will not cause the extra leakage of privacy.

\subsection{Sparse Classification method with Differential Privacy}

In this subsection, we propose differential privacy sparse classification algorithm, and give theoretical privacy bound of the algorithm. 

To achieve differential privacy, 
we perturb the results of each iteration of ADMM by adding noise. 
Notice that the original data will only be accessed when updating variable $w$ at step 2, 
so we perturb the objective function (4) by adding $cb^Tw$ during each iteration, where $b$ is a random vector, with the probability density proportional to $\exp(-\gamma\|b\|_2)$, and $\gamma$ is the privacy parameter which controls the degree of privacy leakage. To generate this noisy vector, we choose the norm from the gamma distribution with shape $p$ and scale $1/\gamma$ and the direction uniformly, where $p$ is the number of the feature.

Specifically, 
in iteration $k+1$, we first update variable $Z$ via (3), then
update variable $w$ by the following perturbing optimization

\begin{align}
w(k+1)&=\underset{w}{\arg\min}\ l_w^{priv}(k+1) \nonumber\\
&=\underset{w}{\arg\min}\ \frac{1}{n}\sum_{i=1}^nO(y_{i}w^Tx_{i})+\frac{c}{2}(\|Z(k+1) -w +\frac{V(k)}{c}\|_2^2)  +cb^Tw, \tag{6}
\end{align}
At last, update Lagrange multiplier $V$ via (5).
Differential privacy sparse classification algorithm(DPSC) is tabulated as Algorithm 1.

\begin{algorithm}
	\caption{Differential Privacy Sparse Classification Algorithm(DPSC)} 
	\label{alg1}
	\begin{algorithmic}
		\STATE{Input: $D=\{x_i,y_i\}_{i=1}^n$, parameter $c$,
			maximum number of iterations $K$, and privacy degree $\epsilon$.} 
		\STATE{Initialize: Generate $Z(0)$,$\ w(0)$ randomly, let $V(0)=\mathrm{0}$.}	
		\STATE{Generate noise $b \sim \exp(-\gamma\|b\|_2)$.}	
		\FOR{$k=0$ to $K-1$}	
		\STATE{\ Update $Z(k+1)$ via (3).}
		\STATE{\ Update $w(k+1)$ via (6).}
		\STATE{\ Update  $V(k+1)$ via (5).}
		\ENDFOR
		\STATE{Output: $w^{*}=w(K)$.}	
	\end{algorithmic}
	
\end{algorithm}

\

We now state our main result on the privacy property of DPSC. 
First, we give the theoretical privacy bound of each iteration of DPSC. 

\

\noindent \textbf{Lemma 2\ } Suppose that Assumption 1-3 hold, and the preselected coefficient $c \geqslant \frac{2c_2}{n}$, where $c_2$ is the upper bound of the second derivative of the loss function $O$, then the iteration $k$ of DPSC satisfies the $\epsilon _{k}-$ DP with
$$ \epsilon _{k}\  =\ （\frac{2\gamma c_1+2.8c_2}{nc},$$
where $\gamma$ is a differential privacy parameter, and $c_1$ is the upper bound of the first derivative of the loss function.

\noindent \textbf{Proof\ }
We prove that for any two data sets $D$ and $D^{'}$ differing on at most one data point, given $\{Z(r)\}_{r=1}^{k-1}$, $\{w(r)\}_{r=1}^{k-1}$ and $\{V(r)\}_{r=1}^{k-1}$,
iteration $k$ of DPSC satisfies
$$\frac{P(\{Z(k),w(k)\}\in S(k)|\ \{Z(r),w(r)\}_{r=1}^{k-1},D)}{P(\{Z(k),w(k)\}\in S(k)|\ \{Z(r),w(r)\}_{r=1}^{k-1},D^{'})}\leq \exp(\epsilon _{k}), \eqno(7)$$
where $S(k)$ is the set of possible outputs in iterations $k$.

We have 
\begin{align}
&\frac{P(\{Z(k),w(k)\}\in S(k)|\ \{Z(r),w(r)\}_{r=1}^{k-1},D)}{P(\{Z(k),w(k)\}\in S(k)|\ \{Z(r),w(r)\}_{r=1}^{k-1},D^{'})} \nonumber\\
&=\displaystyle \frac{P( Z(k)=Z^{*}(k) \ |\ \{Z(r),w(r)\}_{r=1}^{k-1},D)} {P(Z(k)=Z^{*}(k)\ |\ \{Z(r),w(r)\}_{r=1}^{k-1},D^{'})} \cdot \frac{P( w(k)=w^{*}(k)\ |\ \{Z(r),w(r)\}_{r=1}^{k-1},Z^{*}(k),D)} {P( w(k)=w^{*}(k)\ |\ \{Z(r),w(r)\}_{r=1}^{k-1},Z^{*}(k),D^{'})}. \tag{8}
\end{align}
Next, we analyze the two parts of (8) separately. 

For the first part of (8), 
because the update of $Z$ is independent of the data, this step will not leak privacy even though no noise is added, and we have
$$\frac{P( Z(k)=Z^{*}(k)\ |\ \{Z(r),w(r)\}_{r=1}^{k-1},D)} {P( Z(k)=Z^{*}(k) \ | \ \{Z(r),w(r)\}_{r=1}^{k-1},D^{'}) } =1.\eqno(9)$$

For the second part of (8),
by the KKT condition, we have $\nabla_{w(k)}l_w^{priv}(k)=0$, which implies
$$b=-\frac{1}{cn}\sum_{i=1}^{n}y_iO^{'}(y_iw^T(k)x_i)x_i+(Z(k)-w(k)+\frac{V(k-1)}{c}).\eqno(10)$$
Given $\{Z(r),w(r),V(r)\}_{r=1}^{k-1}$ and $Z^{*}(k)$, $B$ and $\mathfrak{W}$ will be bijective, where $B$ and $\mathfrak{W}$ are random variables, of which the realizations are $b$ and $W(k)$. Now, let $g_k(\cdot,D) : \mathbb{R}^p \to \mathbb{R}^p$ denote the one-to-one mapping from $B$ to $\mathfrak{W}$ using dataset $D$, then

\begin{align}
&\frac{P( w(k)=w^{*}(k)\ |\ \{Z(r),w(r)\}_{r=1}^{k-1},Z^{*}(k),D)} {P( w(k)=w^{*}(k) \ | \ \{Z(r),w(r)\}_{r=1}^{k-1},Z^{*}(k),D^{'}) }\nonumber\\
&= \frac{P(g^{-1}_k(w(k),D) )} {P(g^{-1}_k(w(k),D^{'}) )}\cdot \frac{| \det(J(g^{-1}_k(w(k),D)))| }{| \det(J(g^{-1}_k(w(k),D^{'})))|}.\tag{11} 
\end{align}
where $g^{-1}_k(w(k),D)$ is the mapping from $w(k)$ to $b$ using data $D$
and $J(g^{-1}_k(w(k),D)$ is the Jacobian matrix of it.

For the first part in (11),
notice that Assumption 2 holds, and for any data point, $\|x_{i}\|_2 \leq 1$ , $y_{i} \in \{-1,1\} $. 
We choose $b \sim \exp(-\gamma\|b\|_2)$, 
and there is only one data difference between $D$ and $D^{'}$,
say $(x_{1},y_{1})$ and $(x_{1}^{'},y_{1}^{'})$ respectively,
so from (10) we have

$$
\|b^{'}-b\|_2=\frac{1}{cn}\|y_1O^{'}(y_1w^{T}(k)x_1)-y^{'}_1O^{'}(y^{'}_1w^{T}(k)x^{'}_1)\|_2   \leq \frac{2c_1}{cn}.  \eqno(12)$$
Thus, for fixed $\gamma$, we have
\begin{align}
\frac{P(g^{-1}_k(w(k),D) )} {P(g^{-1}_k(w(k),D^{'}) )}&= \exp(\gamma\|b^{'}\|_2-\gamma\|b\|_2)
\nonumber\\ 
& \leq \exp(\gamma\|b^{'}-b\|_2) 
\leq \exp (\frac{2\gamma c_1}{cn}).\tag{13}
\end{align}

Consider the second part in (11), the Jacobian matrix is 
$$
J(g^{-1}_k(w(k),D)=-\frac{1}{cn}\sum_{i=1}^{n}(O^{''}(y_iw^T(k)x_i)x_ix^T_i)
-I_{p\times p},
$$
let $G(k)=\frac{1}{cn}
(O^{''}(y^{'}_1w^T(k)x^{'}_1)x^{'}_1x^{'T}_1-O^{''}(y_1w^T(k)x_1)x_1x^{T}_1)$,
and $H(k)=-J(g^{-1}_k(w(k),D)$, then
\begin{align}
&\frac{| \det(J(g^{-1}_k(w(k),D)))| }{| \det(J(g^{-1}_k(w(k),D^{'})))|}=\frac{|\det(H(k))|}{|\det(H(k)+G(k))|}  \nonumber\\
&=\frac{1}{|\det(I+H(k)^{-1} \cdot G(k))|}=\frac{1}{\displaystyle\prod_{i=1}^{r}(1+\lambda_{i}(H(k)^{-1}\cdot G(k)))}, \tag{14}
\end{align}
where $\lambda_{i}(H(k)^{-1} \cdot G(k)))$ denotes the $i$-th largest eigenvalue of $H(k)^{-1}G(k)$.
Since $G(k)$ has rank at most 2, it implies that $H(k)^{-1}G(k)$ also has rank at most 2.
Notice that $0< O^{''} \leq c_2,$
thus
$\lambda_i(H(k)) \geq 1 > 0,$ and $ -\frac{c_2}{cn} \leq \lambda_i(G(k)) \leq  \frac{c_2}{cn},$
which implies that 
$$ -\frac{c_2}{cn} \leq \lambda_j(H(k)^{-1}G(k)) \leq  \frac{c_2}{cn}. $$
By the assumption $c > \frac{2c_2}{n}$,
we have $$ -\frac{1}{2} \leq \lambda_j(H(k)^{-1}G(k)) \leq \frac{1}{2},$$
since $ \lambda_{min}(H(k)^{-1}G(k)) \geq -1,$ then
$$
\frac{1}{|1+\lambda_{max}(H(k)^{-1}G(k))|^2} \leq \frac{1}{|\det(I+H(k)^{-1}G(k))|}  \leq \frac{1}{|1+\lambda_{min}(H(k)^{-1}G(k))|^2}.$$
thus
$$
\frac{| \det(J(g^{-1}_k(w(k),D)))| }{| \det(J(g^{-1}_k(w(k),D^{'})))|}  \leq \frac{1}{|1-\frac{c_2}{cn}|^2}\leq \exp(-2\ln(1-\frac{c_2}{cn}))
\leq \exp{(\frac{2.8c_2}{nc})}.\eqno(15)$$
where in the last inequality, we use the fact that for any $x \in [\ 0,\ \frac{1}{2}\ ],$ $-\ln(1-x)<1.4x.$
Combining (13) and (15),
$$\frac{P( w(k)=w^{*}(k)\ |\ \{Z(r),w(r)\}_{r=1}^{k-1},Z^{*}(k),D)} {P( w(k)=w^{*}(k) \ | \ \{Z(r),w(r)\}_{r=1}^{k-1},Z^{*}(k), D^{'}) }  \leq \exp (\frac{2\gamma c_1+2.8c_2}{cn}).\eqno(16)$$

Combining (9) and (16) , 
the lemma has been proved.

\

\noindent \textbf{Theorem 3\ } Suppose that Assumption 1-3 hold, and the preselected coefficient $c \geq \frac{2c_2}{n}$, then DPSC satisfies the $\epsilon-$ differential privacy with $\epsilon\ =\ K\cdot (\frac{2\gamma c_1+2.8c_1}{cn})$, where $K$ is the number of iteration.

\noindent \textbf{Proof\ }
We prove that for any two datasets $D$ and $D^{'}$ that differing on at most one data point, 
DPSC satisfies
$$\frac{P(\{S(k)\}_{k=1}^K\in S|D)}{P(\{S(k)\}_{k=1}^K\in S|D^{'})}\leq \exp(\epsilon).$$
where $S$ is the set of possible outputs during $K$ iterations.

We have 
\begin{align}
&\frac{P( \{Z(k),w(k)\}_{k=1}^K \in S\ |\ D)} {P( \{Z(k),w(k)\}_{k=1}^K \in S\ |\ D^{'})} \nonumber\\
&= \frac{P( \{Z(0),w(0)\} \in S(0)\ |\ D)} {P( \{Z(0),w(0)\} \in S(0)\ |\ D^{'})}  \cdot \displaystyle\prod_{k=1}^{K}\frac{P( \{Z(k),w(k)\} \in S(k)\ |\  \{Z(r),w(r)\}_{r=1}^{k-1},D)} {P( \{Z(k),w(k)\} \in S(k)\ |\ \{Z(r),w(r)\}_{r=1}^{k-1},D^{'})}  \nonumber\\
&= \displaystyle\prod_{k=1}^{K}\frac{P( \{Z(k),w(k)\} \in S(k)\ |\  \{Z(r),w(r)\}_{r=1}^{k-1},D)} {P( \{Z(k),w(k)\} \in S(k)\ |\ \{Z(r),w(r)\}_{r=1}^{k-1},D^{'})},\tag{17}
\end{align}
and since we generate $ Z(0)$ and $w(0)$ randomly, the last equality achieved.
Lemma 2 shows that for each $k$, $k\ge 1$, 
$$\frac{P(\{Z(k),w(k)\}\in S(k)|\ \{Z(r),w(r)\}_{r=1}^{k-1},D)}{P(\{Z(k),w(k)\}\in S(k)|\ \{Z(r),w(r)\}_{r=1}^{k-1},D^{'})}\leq \exp(\frac{2\gamma+2.8c_1}{cn}), 
$$
Thus we have
$$\frac{P(\{S(k)\}_{k=1}^K\in S|D)}{P(\{S(k)\}_{k=1}^K\in S|D^{'})}\leq \exp(K\cdot(\frac{2\gamma+2.8c_1}{cn})).$$
Hence the theorem is proved.

\

From Theorem 3, we can see that the privacy bound of DPSC is controled by the iteration number $K$, the privacy parameter $\gamma$, the upper bound of the first and  second derivative of loss function $c_1$ and $c_2$, ADMM pre-selected parameter $c$, and the data size $n$. 
With increasing of $\gamma$, 
the privacy bound increase, 
which means that the algorithm leaks more privacy. 
Meanwhile, the noisy vector $b$ decrease, 
that is, we perturb the objective function with a smaller scale. 
Thus, the utility of algorithm increase. 
In addition, as the size of training data increase, the privacy bound decrease, i.e. the possibility of individual privacy leakage decrease. 
It shows that the large traning data make individual more difficult to be distinguished. 

\

\noindent \textbf{Remark 3\ }
Based on the idea of perturbing the optimization problem\cite{Chaudhuri}, we consider noising the objective function of $Z$ and $w$ in each itertaion of ADMM to achieve differential privacy. Notice that the update of $Z$ is independent of the data,  that is, the original data will not be accessed when updating variable $Z$. Thus we only perturb the objective function of $w$, then propose DPSC algorithm.

\section{Applications}

In this section, we apply the differential privacy sparse classification model to two typical models - logistic regression with $L_1$ regularizer and logistic regression with $L_{1/2}$ regularizer. We propose privacy classification algorithms, and present the privacy bound of them.
The applications show that our method is suitable for dealing with sparse classification problems.

\subsection{Differential Privacy Logistic Regression with $L_1$ Regularizer}

The logistic regression with $L_1$ regularizer \cite{Park} is a useful method to solve over-fitting problem and has well  generalization \cite{Genkin,Koh}. It selects variables according to the amount of penalization on the $l_1$-norm of the coefficients. 
The logistic regression with $L_1$ regularizer has the following form
$$\underset{w}{\min}\ \frac{1}{n}\sum_{i=1}^n\log(1+\exp(-y_{i}w^Tx_{i}))+\lambda\|w\|_1,\eqno(18)$$
where $x_i\in \textbf{R}^p$ is input variable, $y_i\in\{-1,1\}$ is output variable. First, we add the auxiliary variable $Z$, and rewrite the model as 
$$
\underset{w,Z}{\min}\ \frac{1}{n}\sum_{i=1}^{n}\log(1+\exp(-y_{i}w^Tx_{i}) )+\lambda\|Z\|_1
$$
$$s.t.\ w=Z.\eqno(19)$$
which has the quadratically augmented Lagrangian function
$$ L(Z,w,V)=\frac{1}{n}\sum_{i=1}^n\log(1+\exp(-y_{i}w^Tx_{i}) )+\lambda\|Z\|_1$$ $$\ \ \ \ \ \ +\frac{c}{2}(\|Z-w+\frac{V}{c}\|_2^2-\|\frac{V}{c}\|_2^2), $$
where $V$ is Lagrange multipliers. From Algorithm 1, Differential Privacy Logistic Regression with $L_1$ Regularizer algorithm (DPLL) has the following steps

\

\noindent Step 1: $Z$ updates,
$$
Z(k+1)=\underset{Z}{\arg\min}\ \lambda\|Z\|_1+\frac{c}{2}(\|Z-w(k)+\frac{V(k)}{c}\|_2^2), \eqno(20)
$$

\noindent Step 2: $w$ updates,
\begin{align}
w(k+1)=&\underset{w}{\arg\min}\ \frac{1}{n}\sum_{i=1}^n\log(1+\exp(-y_{i}w^Tx_{i}) ) \nonumber\\
&+\frac{c}{2}(\|Z(k+1)-w+\frac{V(k)}{c}\|_2^2)+cb^Tw, \tag{21}
\end{align}

\noindent Step 3:\ Lagrange multiplier $V$ updates,
$$V(k+1)= V(k)+c(Z(k+1)-w(k+1)).\eqno(22)$$

Consider the updating of $Z$, let $Q(k)=w(k)-\frac{V(k)}{c}$, for $i=1,\cdots,p$, the coordinate $Z_i$ of vector $Z$ is updated by

$$Z_i(k+1)=\underset{Z_i}{\arg\min}\ \lambda|Z_i|+\frac{c}{2}(Q_{i}(k)-Z_i)^2,\eqno(23)$$
where $Q_{i}(k)$ is the $i$th coordinate of vector $Q(k)$.
It's easy to obtain the solutions of (23), which are
$$ Z_i(k+1)=\left\{
\begin{array}{rcl}
Q_{i}(k)-\frac{\lambda}{c}     &      & {if \  Q_{i}(k)\ge \frac{\lambda}{c} } \\
0 \ \ \ \ \ \ \ \ \ \ \ \ \ &  & {if \ -\frac{\lambda}{c} < Q_{i}(k)\leq \frac{\lambda}{c} }\\
Q_{i}(k)+\frac{\lambda}{c}     &      & { if \ Q_{i}(k) < \frac{\lambda}{c}.  \  }
\end{array} \right. \eqno(24)$$

To update variable $w$, we consider using gradient descent algorithm.  
Let $m=1,2,\cdots,M$ denote the inner iteration index for the gradient descent algorithm used to solve (21).
For the minimization at step $k+1$ of the (outer) consensus iteration,
the sequence of iterates $w^0(k+1)=w(k)$, at step $m$
$$w^m(k+1)=w^{(m-1)}(k+1)-\alpha\frac{\partial l_w^{priv}(w^{(m-1)}(k+1))}{\partial w},\eqno(25)$$
where $\frac{\partial l_w^{priv}(w^{(m-1)}(k+1))}{\partial w}=
\frac{1}{n}\sum_{i=1}^{n}\frac{-y_{i}x_{i}}{1+\exp(y_{i}w^Tx_{i})}-c(Z(k+1)-w^{(m-1)}(k)+\frac{V(k)}{c})+c\cdot b,$ and $\alpha$ is learning rate.

Now, we present the Differential Privacy Logistic regression with $L_1$ regularizer algorithm (DPLL). The overall operation of the algorithms can be described as follows. 
During iteration $k+1$, we update each coordinate $i$ of $Z$ via (24), then update $w$ cyclically via (25), at last, update Lagrange multiplier $V$ via (22).

\begin{algorithm}
	\caption{Differential Privacy Logistic Regression with $L_1$ Regularizer algorithm(DPLL)} 
	\label{alg2}
	\begin{algorithmic}
		\STATE{Input: $D=\{x_i,y_i\}_{i=1}^n$, parameter $c$,
			maximum number of iterations $K$,  maximum number of  inner iteration $M$,
			learning rate $\alpha$, and privacy degree $\epsilon$.}
		\STATE{Initialize: Generate $Z(0)$ and $ w(0)$ randomly, and let $V(0)=(0,...,0)$.}	
		\STATE{Generate noise $b \sim \exp(-\gamma\|b\|_2)$.}	
		\FOR{$k=0$ to $K-1$}				
		\FOR{$i=1$ to $p$}
		\STATE{ Update $ Z_i(k+1)$ via (24).			
		}		
		\ENDFOR	
		\FOR{$m=1$ to $M$}
		\STATE{Update $w^m(k+1)$ via (25).}
		\ENDFOR
		\STATE{Let $w(k+1)=w^M(k+1)$.}
		\STATE{Update $V(k+1)$ via (22).}
		\ENDFOR
		\STATE{Output: $w^{*}=w(K)$.}	
	\end{algorithmic}
	
\end{algorithm}

The theoretical privacy bound of DPLL is presented in next subsection.

\subsection{Differential Privacy Logistic Regression with $L_{1/2}$ Regularizer}

In this subsection, we focus on logistic regression with nonconvex penalty. As a typical nonconvex regularization, $L_{1/2}$ has many desirable properties and has been studied extensively. 
\cite{zhao1} proposed the logistic regression with $L_{1/2}$ regularizer
which enhances the variable selection capability and alleviates the over-fitting problem of the traditional model. 
It has the following form
$$\underset{w}{\min}\ \frac{1}{n}\sum_{i=1}^n\log(1+\exp(-y_{i}w^Tx_{i}))+\lambda\|w\|_{1/2}^{1/2},\eqno(26)$$
where $\|w\|_{1/2}^{1/2}=\sum_{i=1}^{p}|w_i|^{1/2}$ is quasi-norm. 
Same as before, we add auxiliary variable $Z$, and rewrite the augmented Lagrangian function as 
$$ L(Z,w,V)=\frac{1}{n}\sum_{i=1}^n\log(1+\exp(-y_{i}w^Tx_{i}) )+\lambda\|Z\|_{1/2}^{1/2}$$ $$\ \ \ \ \ \ +\frac{c}{2}(\|Z-w+\frac{V}{c}\|_2^2-\|\frac{V}{c}\|_2^2). $$
Based on Algorithm 1, we propose Differential Privacy Logistic regression with $L_{1/2}$ regularizer (DPLH), which is tabulated as Algorithm 3. Here, the update of $w$ and $V$ in each iteration are same as DPLL.

\begin{algorithm}
	\caption{Differential Privacy Logistic Regression with $L_{1/2}$ Regularizer (DPLH)} 
	\label{alg3}
	\begin{algorithmic}
		\STATE{Input: $D=\{x_i,y_i\}_{i=1}^n$, parameter $c$,
			maximum number of iterations $K$, maximum number of  inner iteration $M$,
			learning rate $\alpha$, and privacy degree $\epsilon$.}
		\STATE{Initialize: let  $Z(0)=(1,...,1)\ $, $ w(0)=1,...,1$, and $V(0)=(0,...,0)$.}	
		\STATE{Generate noise $b \sim \exp(-\gamma\|b\|_2)$.}	
		\FOR{$k=0$ to $K-1$}
		
		\STATE{Update $Z(k+1)$ via}  
		\STATE{   	$Z(k+1)=\underset{Z}{\arg\min}\ \lambda\|Z\|_{1/2}^{1/2}+\frac{c}{2}(\|Z-w(k)+\frac{V(k)}{c}\|_2^2.\ \ (27)$ }				
		\FOR{$m=1$ to $M$}
		\STATE{Update $w^m(k+1)$ via (25).}
		\ENDFOR
		\STATE{Let $w(k+1)=w^M(k+1)$.}
		\STATE{Update $V(k+1)$ via (22).}
		
		\STATE{Output: $w^{*}=w(K)$.}
		\ENDFOR		
	\end{algorithmic}	
\end{algorithm}

\

In iteration $k+1$ of DPLH, to solve (27), we resort to reweighted algorithm, which has the following steps

\

\noindent Step 1: Set the initial value $Z^0$ and the maximum iteration step $T$, and let $t = 0$.

\

\noindent Step 2: Solve
$$  Z^{t+1}= \underset{Z}{\arg\min}\ \frac{c}{2}(\|Z-w(k)+\frac{V(k)}{c}\|_2^2)+\lambda\sum_{i=1}^{p}\frac{|Z_i|}{|Z_i^t|^{1/2}}$$
with an existing $L_1$ algorithm, and let $t=t+1.$

\

\noindent Step 3: If $t < T$, go to Step 2, otherwise, output $Z(k+1)=Z^t$.

\

\noindent The initial value $Z^0$ normally can be taken as $Z^0 = (1, 1, . . . , 1)$. And with such a setting, the first iteration ($t = 0$) in Step 2 is exactly corresponding to solving a lasso problem. When $t = 1$, with easy linear transformation, step 2 can be transformed into a lasso problem.
It is possible that when $t>1$, some $Z^t_i=0$. To guarantee the feasibly, we replace $|Z_i^t|^{1/2}$ with $|Z_i^t+\mu|^{1/2}$ in Step 2, where $\mu$ is any fixed positive real number.

\

The overall operation of DPLH can be described as follows. 
During iteration $k+1$, we update variable $Z$  via reweighted algorithm, then update $w$ cyclically via (25), at last, update Lagrange multiplier $V$ via (22).

\

Now, we present the theoretical privacy bound of the DPLL and DPLH.

\

\noindent \textbf{Theorem 4\ }
Suppose that preselected penalty coefficient $c \ge \frac{1}{2n}$ and $\gamma \leq cn-\frac{7}{20}$, then DPLL and DPLH satisfy the $\epsilon-$ differential privacy with $ \epsilon\ =\  K\cdot \epsilon _1,$
where $ \epsilon _1\  =\ \frac{8\gamma+2.8}{4cn}$, and $K$ is the iteration number.

\noindent \textbf{Proof\ }
Obviously, the logistic	function is strongly convex, Lipschitz differentiable, with $0\leq O^{''}\leq \frac{1}{4}$, which satisfies Assumption 2, here $c_1=\frac{1}{4}$. And $L_1$ regularizer and $L_{1/2}$ regularizer satisfy Assumption 3. Thus, from the proof of Theorem 3, theorem is established.

\section{ Experiments}

We use the simulation and real data to illustrate the effectiveness of our algorithms DPLL and DPLH. In all datasets, we preprocessed the data such that the input vectors had maximum norm 1.

\subsection{ Simulation}

We study the classification model  $$p(y_i=1|x_i)=\frac{1}{1+\exp(-w^Tx_i)},\ i=1,\dots,n.$$
where $x_i\in \mathbf{R}^{100} $ is an input vector, 
$x_{i}\sim N(0,\Sigma),$
covariance matrix $\Sigma=0.5^{|i-j|},\ 1\leq i,j \leq n,$
coefficient vector $w\in \mathbf{R}^{100}$, where 
$w (1:8)= (10,9,8,7,6,5,4,0.5)'$, and $w(9:100)=(0,0,...,0)$.
$\epsilon$ is standardized normal random error, 
and sample number $n$. 
If $p(y_i=1|x_i)\geq 0.5,$ let $y_i=1$, else, $y_i=-1.$

\subsubsection{ Privacy-Accuracy Tradeoff }

\

In this case, we study the tradeoff between the privacy requirement on the classifier, and its classification accuracy, when the classifier is trained on data of a fixed size. 
The privacy bound of DPLL and DPLH is $ \epsilon\ =\  K\cdot \ \frac{8\gamma+2.8}{4cn}$. Thus, the privacy requirement  is measured by the privacy parameter $\gamma$, iteration number $K$, pre-selected parameter $c$ and training data size $n$. In this case, we set $K=100$, $c=2.5$, and $n=10000$.   
Then  we run DPLL and DPLH with $\gamma$ increases. To measure accuary of two algorithms,  we record the classification error rate (CE) and mean square error (MSE). We also record the average number of correctly identified zero coeffcients (C.A.N of zero) and the average number of incorrectly identfied zero coe cients (IC.A.N of zero) over the datasets to measure the ability of variable selection of two algorithms.

We have simulated 50 datasets consisting of 11,000 observations， and each data set was divided into two parts: a training set with 10,000 observations and a test set with 1000 observations. To demonstrate the effectiveness of DPLL and DPLH, we applied the logistic regression with $L_1$ algorithm (denote as LLA), the logistic regression with $L_{1/2}$ algorithm (denote as LHA), DPLL and DPLH to the 50 datasets.
For each data set, we trained classifiers for 5 fixed values of $\lambda$ and tested the error of these classifiers. 
Specifically, for non-private algorithms, we chose the tunning parameter $\lambda$ by 5-fold CV. 
And for private algorithms, we set the same value of $\lambda$ with non-private algorithms for the same dataset. 
We take differential privacy degree $\epsilon$ as $0.1,0.5,1,1.5,2,2.5,3$ and $4$, and let gradient descent step $\alpha$ as 0.5. 
Due to the randomness of the noise, we performed 50 independent runs of two privacy algorithms for each parameter setting, and recorded the average results of 50 trials. 
Figure.1 and Figure.2 show the variable selection results of non-private algorithms and private algorithms.  
Figure.3 and Figure.4 show the classification results of four algorithms.

\begin{figure}[!t]
	\centering
	\begin{minipage}[c]{0.48\textwidth}
		\centering
		\includegraphics[height=4cm,width=7cm,angle=0]{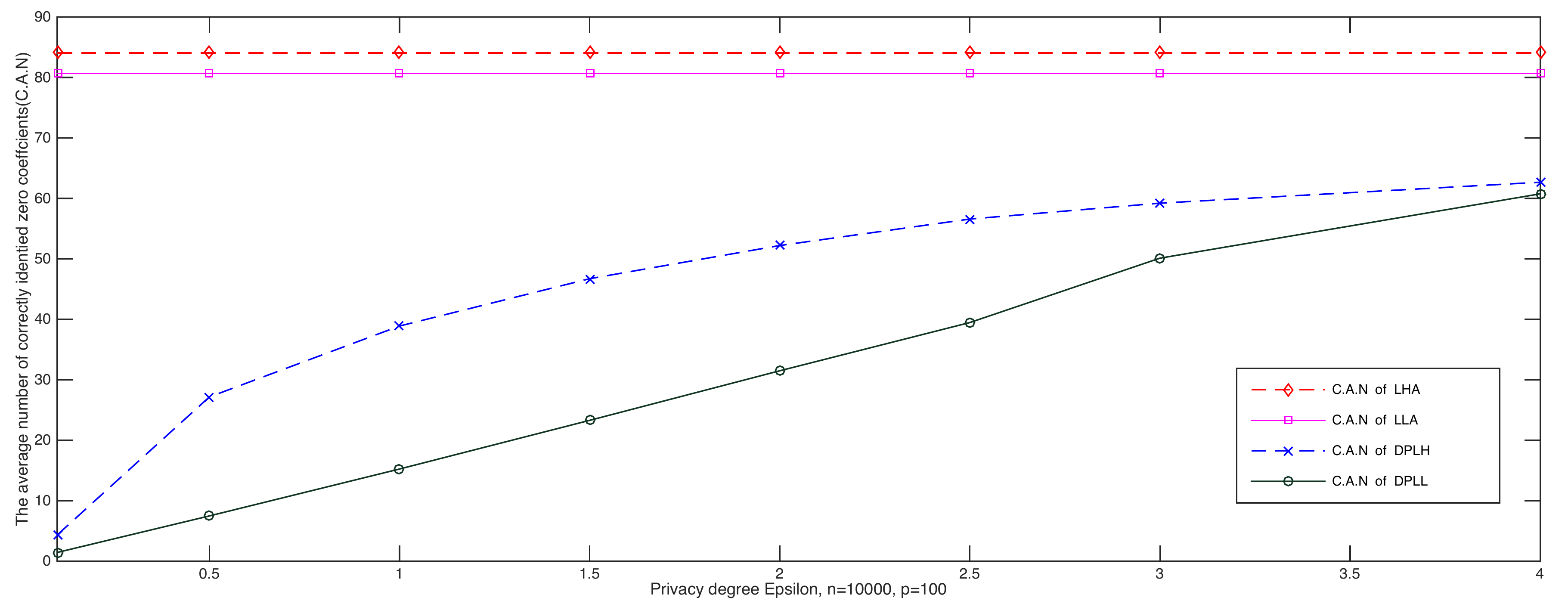}
	\end{minipage}
	\hspace{0.02\textwidth}
	\begin{minipage}[c]{0.48\textwidth}
		\centering
		\includegraphics[height=4cm,width=7cm,angle=0]{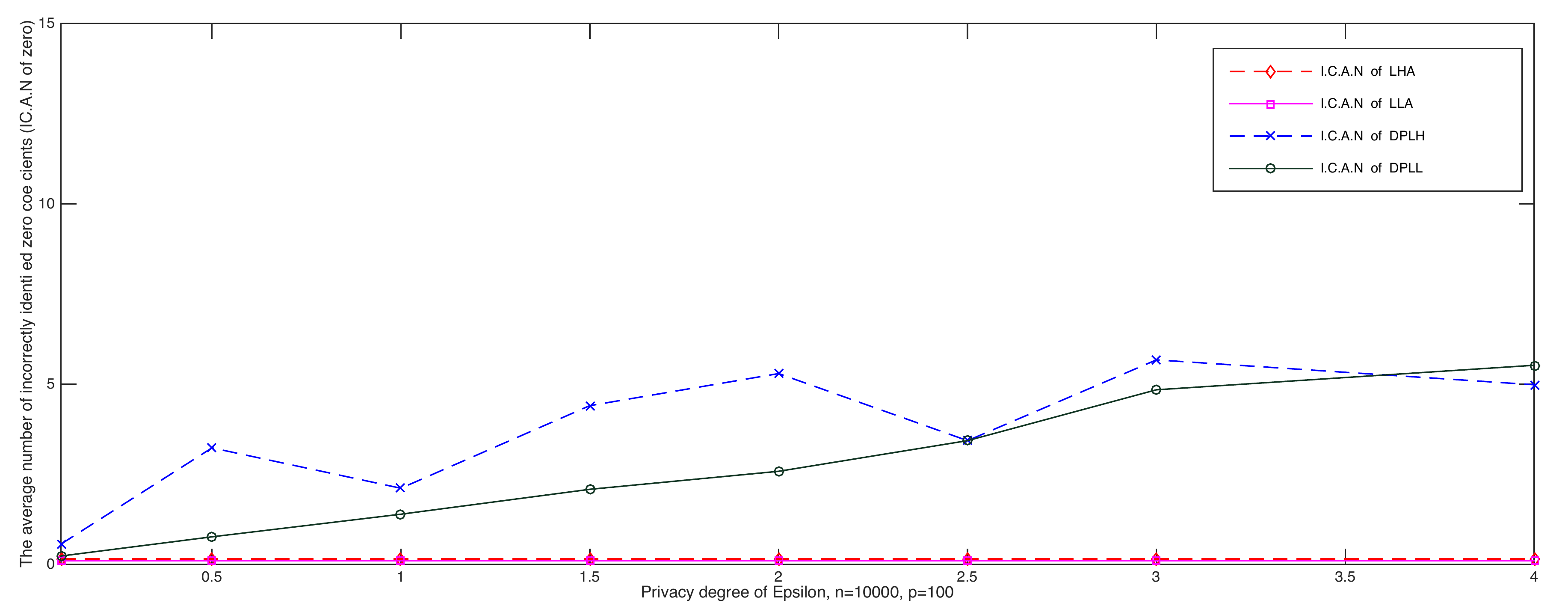}
	\end{minipage}\\[3mm]
	\begin{minipage}[t]{0.48\textwidth}
		\centering
		\caption{With the increase of privacy degree $\epsilon$, results of C.A.N of LLA, LHA,DPLL and DPLH.}
		\label{fig5}
	\end{minipage}
	\hspace{0.02\textwidth}
	\begin{minipage}[t]{0.48\textwidth}
		\centering
		\caption{With the increase of privacy degree $\epsilon$, results of IC.A.N of LLA, LHA,DPLL and DPLH.}
	\end{minipage}
\end{figure}

\begin{figure}[!t]
	\centering
	\begin{minipage}[c]{0.48\textwidth}
		\centering
		\includegraphics[height=4cm,width=7cm,angle=0]{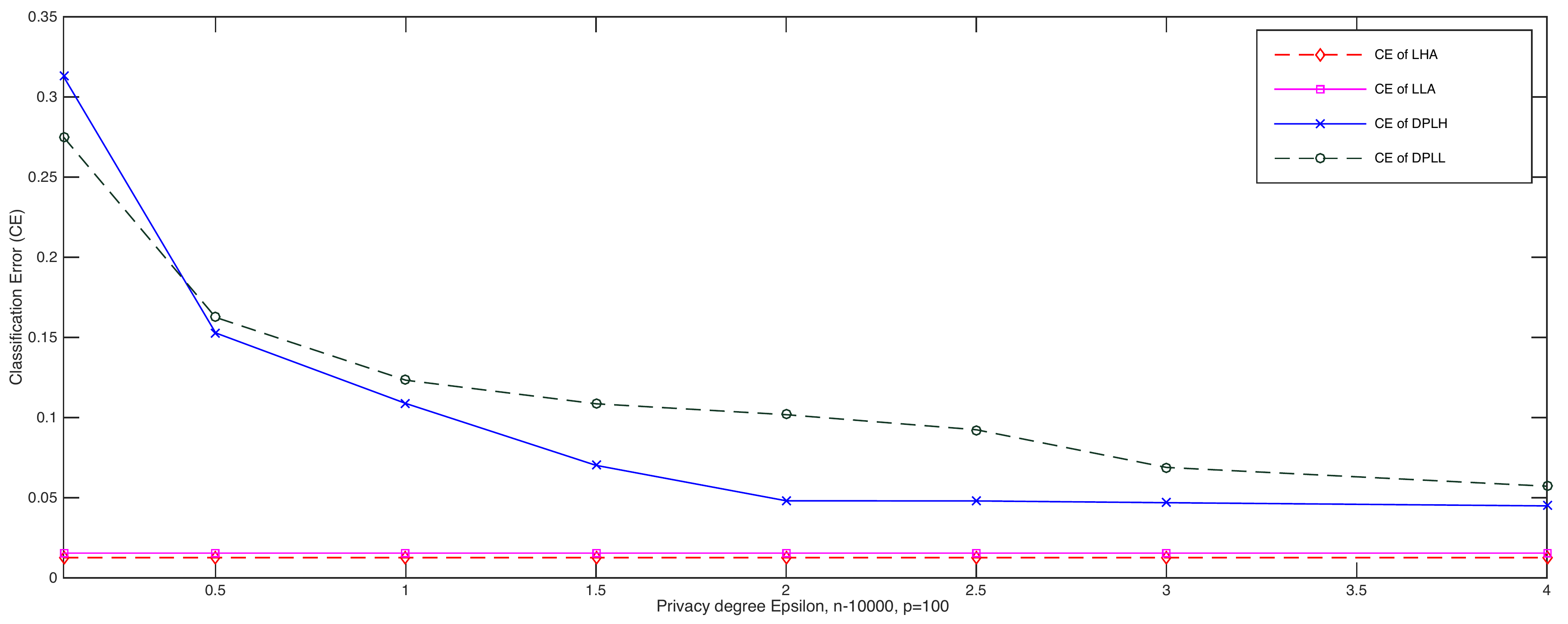}
	\end{minipage}
	\hspace{0.02\textwidth}
	\begin{minipage}[c]{0.48\textwidth}
		\centering
		\includegraphics[height=4cm,width=7cm,angle=0]{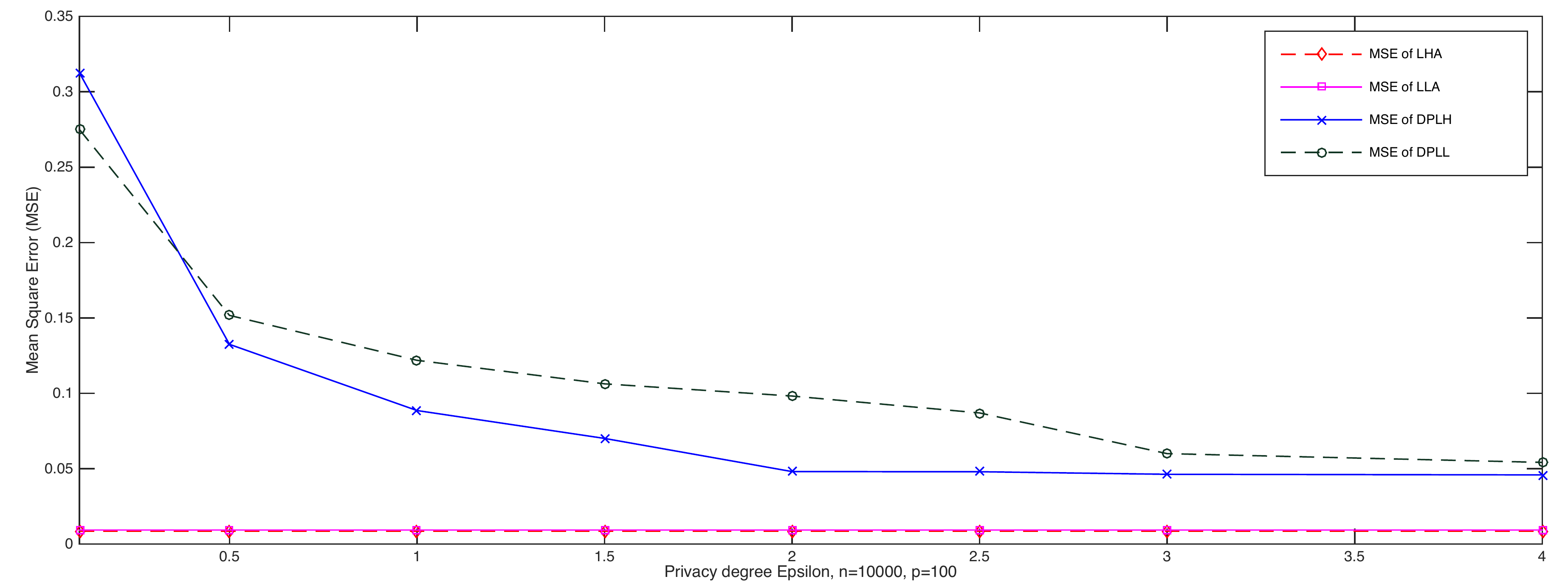}
	\end{minipage}\\[3mm]
	\begin{minipage}[t]{0.48\textwidth}
		\centering
		\caption{With the increase of privacy degree $\epsilon$, results of Classification Error Rate of LLA, LHA,DPLL and DPLH.}
		\label{fig5}
	\end{minipage}
	\hspace{0.02\textwidth}
	\begin{minipage}[t]{0.48\textwidth}
		\centering
		\caption{With the increase of privacy degree $\epsilon$, results of Mean Square Error of LLA, LHA,DPLL and DPLH.}
	\end{minipage}
\end{figure}

From Figure 1 and Figure 2, we can see that C.A.N of zero of LLA and LHA are larger than C.A.N of zero of DPLL and DPLH. And with the increase of privacy degree $\epsilon$, C.A.N of zero of DPLL and DPLH both increase. 
The stricter privacy-preserving (smaller $\epsilon$), the lower variable selection accuracy (smaller C.A.N of zero). 
These results show that privacy and variable selection accuracy can not be simultaneously satisfied when training data size $n$ is fixed and not large enough.  
What's more, the variable selection results of logistic regression with $L_{1/2}$ regularizer are more sparse than logistic regression with $L_{1}$ regularizer, even though IC.A.N of zero of logistic regression with $L_{1/2}$ regularizer is slightly larger than logistic regression with $L_{1}$ regularizer.

From Figure 3 and Figure 4, we can see that prediction error (MSE and CE) of DPLL and DPLH are larger than  prediction error (MSE and CE) of LLA and LHA. And with the increase of privacy degree $\epsilon$,  prediction error (MSE and CE) of DPLL and DPLH both decrease. 
The stricter privacy-preserving (smaller $\epsilon$), the lower prediction accuracy (larger MSE and CE ).
These results show that privacy and prediction accuracy can not be simultaneously satisfied, when training data size $n$ is fied and not large enough.

\subsubsection{ Accuracy-Training data size Tradeoff }

\

In this case, we study the tradeoff between training data size and prediction accuracy. Specifically, we examine how prediction accuracy varies as the size of the training set increases when the privacy degree is fixed.

In this case, we set $K=150$, $c=2.5$, and $\epsilon=0.5$ and $1.5$ separately. 
We have simulated 6 datasets consisting of $5000,10000,...,30000$ training observations， and have simulated a test set with 1000 observations. 
Then we run LLA,  LHA, DPLL and DPLH with training data size $n$ increases. To measure accuary of two algorithms,  we record the classification error rate (CE) and mean square error (MSE). 
For each data set, we also trained classifiers for 5 fixed values of $\lambda$, and let gradient descent step $\alpha$ as 0.5. 
Due to the randomness of the noise, we performed 50 independent runs of two privacy algorithms for each parameter setting, and recorded the average results of 50 trials.   
Figure.5 and Figure.6 show the prediction results (MSE and CE) of four algorithms.

\begin{figure}[!t]
	\centering
	\begin{minipage}[c]{0.48\textwidth}
		\centering
		\includegraphics[height=4cm,width=7cm,angle=0]{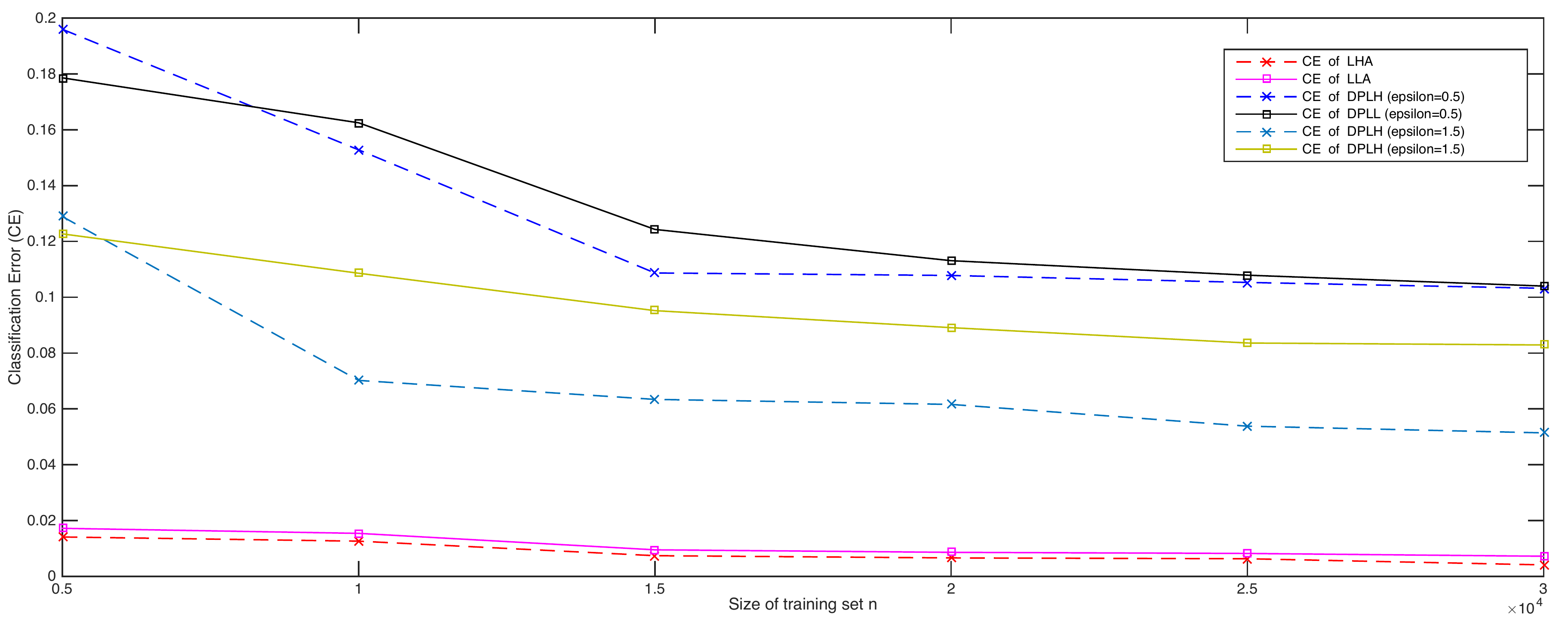}
	\end{minipage}
	\hspace{0.02\textwidth}
	\begin{minipage}[c]{0.48\textwidth}
		\centering
		\includegraphics[height=4cm,width=7cm,angle=0]{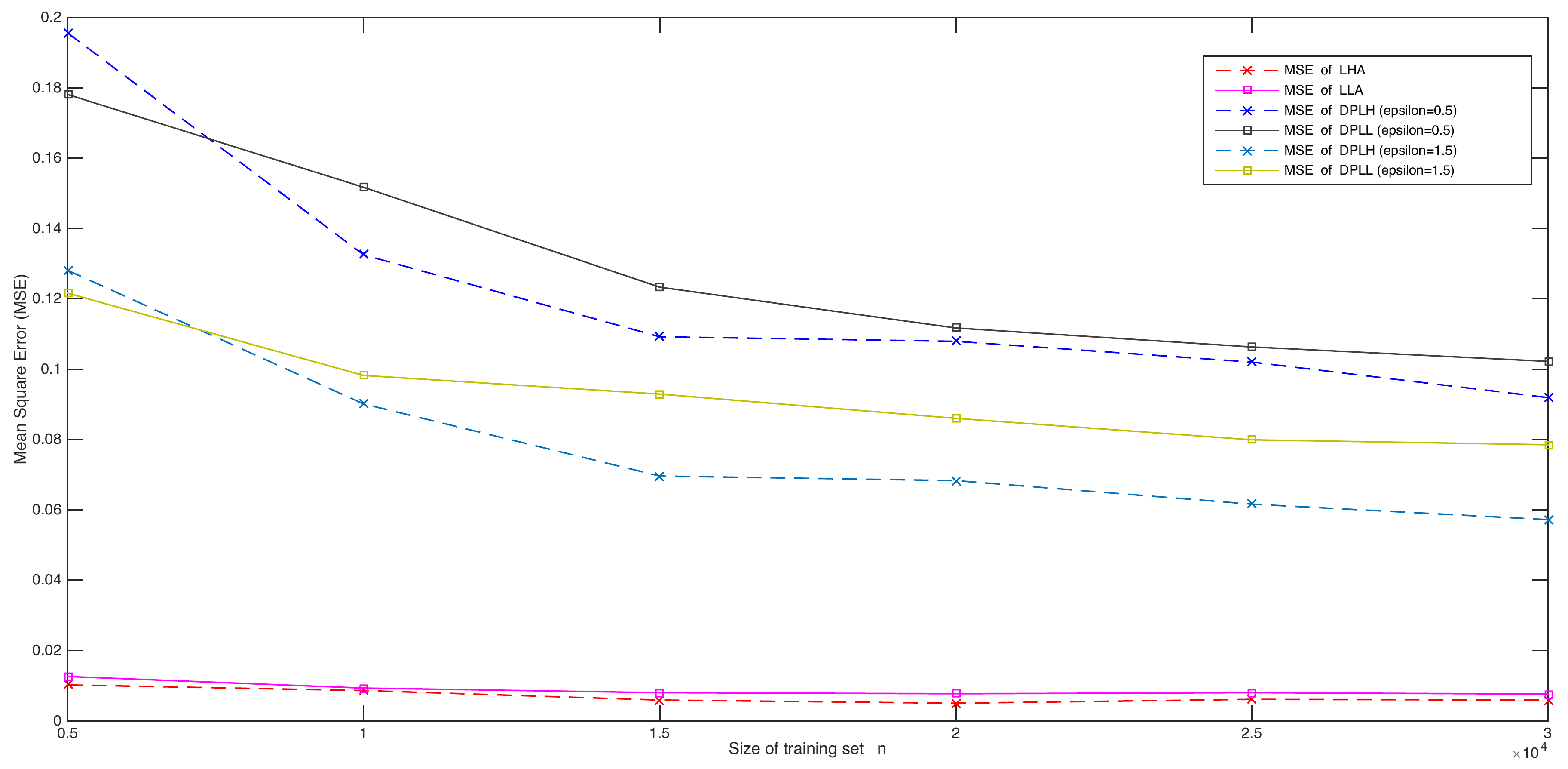}
	\end{minipage}\\[3mm]
	\begin{minipage}[t]{0.48\textwidth}
		\centering
		\caption{With the increase of training data size $n$, results of Classification Error Rate of LLA, LHA,DPLL and DPLH.}
		\label{fig5}
	\end{minipage}
	\hspace{0.02\textwidth}
	\begin{minipage}[t]{0.48\textwidth}
		\centering
		\caption{With the increase of training data size $n$, results of Mean Square Error of LLA, LHA,DPLL and DPLH.}
	\end{minipage}
\end{figure}

From Figure 5 and Figure 6, we can see that prediction error (MSE and CE) of DPLL and DPLH are larger than  prediction error (MSE and CE) of LLA and LHA. And with the increase of the size of training data,  prediction error (MSE and CE) of DPLL and DPLH both decrease for two $\epsilon$. 
The privacy bound $ \epsilon\ =\  K\cdot \ \frac{8\gamma+2.8}{4cn}$ of DPLL and DPLH shows that for fixed $\epsilon$ and other parameters, with the increase of $n$, $\gamma$  which controls the norm of random variable $b$ increases. Thus the prediction error decreases. 
And our experiment results are consistent with this conclusion. 
Interestingly, when the size of training data $n$ is large enough, 
the results of the differential privacy algorithms are similar as the corresponding non-private algorithms.
Big data offers more effectively protection to individual privacy.

\subsection{ Real data}

We analyze the KDDCup99 data set from the UCI Machine Learning Repository  \cite{Hettich}. The task is to build a network classifier, which can predict whether a network connection is a denial-of-service attack or not. 
This data set contains about 5,000,000 instances, and each instance includes
41 attributes. For this data the average fraction of negative labels is 0.20.
To preprocess the data, first we converted each classiffication attribute to a binary vector,
and converted lables $\{good\ connections,$
$\ bad\ connections\}$ to $\{1,-1\}$. 
Then we normalized each column to ensure that the maximum value is 1, and finally, we normalized each row to ensure that the norm of any example is at most 1.
After preprocessing, each instance was represented by a 118-dimensional vector, of norm at most 1.

In this case, we study the privacy-accuracy tradeoff and the accuracy-training data size tradeoff.  We applied LLA, LHA, DPLL and DPLH to the subset of KDDCup99 data set with different parameter settings. 
For non-private and private algorithms, we chose the tunning parameter $\lambda$ by 5-fold CV. 
When applied classification algorithms, let gradient descent step $\alpha$ 
as 0.2, iteration number $K$ as $150$, and pre-selected parameter $c$ as $1$.

To study the privacy-accuracy tradeoff, we chose 60000 examples from the original data set, of which the number of positive label is 38420. We applied four algorithms to this data set, set $\epsilon$ as $0.1,0.5,1,1.5,2,2.5,$ and $3$, 
and performed 50 independent runs of two private algorithms for each parameter setting. We recorded the average results of 50 trials. 
Figure.7 and Figure.8 show the prediction results (MSE and CE) of four algorithms. 

To study the accuracy-training data size tradeoff, 
we chose 6 sub-datasets of size $n=10000$ to  $n=60000$ from the original data, and applied LLA, LHA, DPLL and DPLH to these datasets. 
We also performed 50 independent runs of two private algorithms for each parameter setting, and recorded the average results of 50 trials.    
Figure.9 and Figure.10 show the prediction results. 

From Figure 7-8, we can see that with the increase of the privacy budget, the prediction error of private algorithms decrease. 
From Figure 9-10, we can see that for private algorithms, the prediction error of private algorithms decrease. 
The results are consistent with theoratical results in Theorem 4.
This case shows that our methods are suitable for dealing  with real data.

\begin{figure}[!t]
	\centering
	\begin{minipage}[c]{0.48\textwidth}
		\centering
		\includegraphics[height=4cm,width=7cm,angle=0]{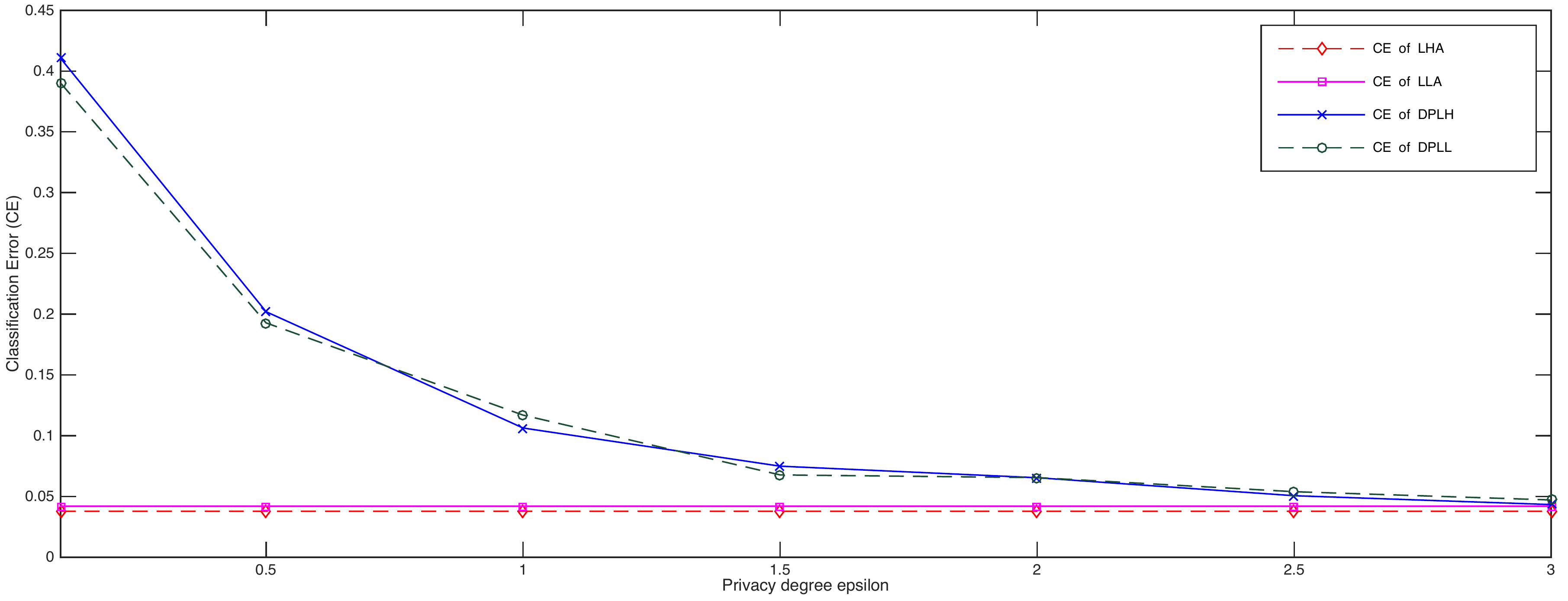}
	\end{minipage}
	\hspace{0.02\textwidth}
	\begin{minipage}[c]{0.48\textwidth}
		\centering
		\includegraphics[height=4cm,width=7cm,angle=0]{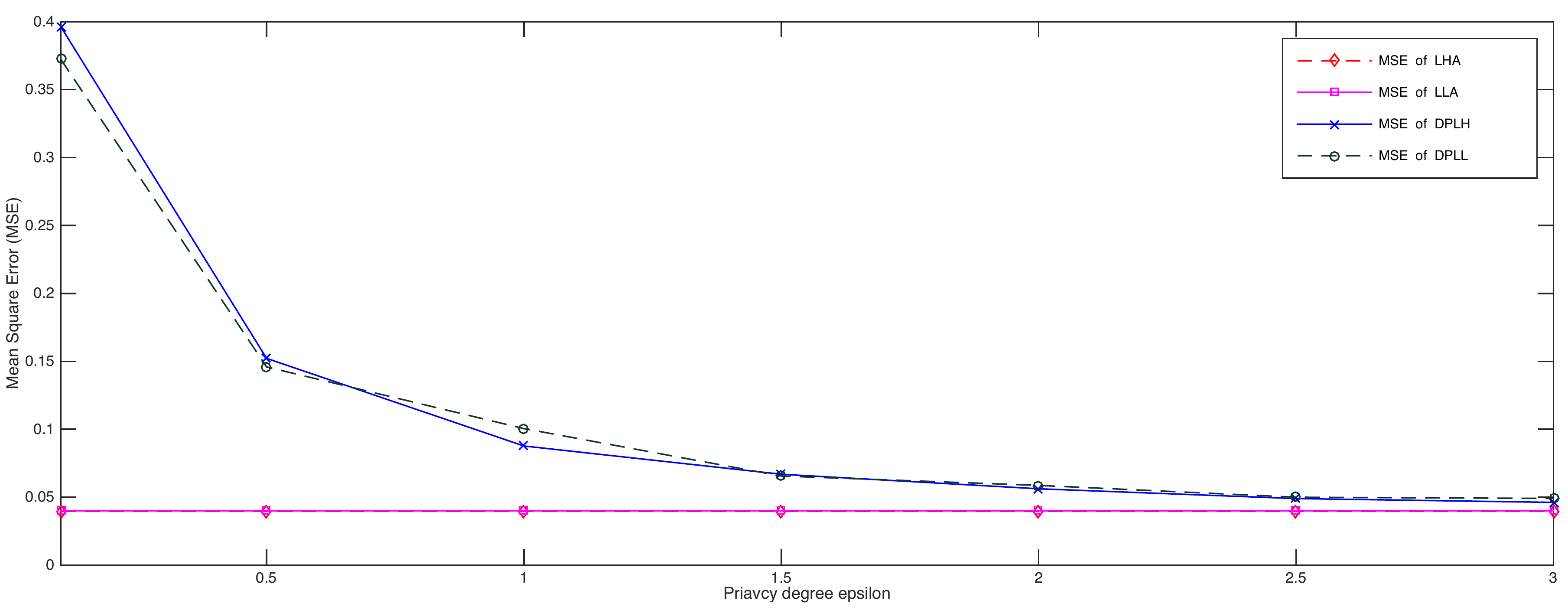}
	\end{minipage}\\[3mm]
	\begin{minipage}[t]{0.48\textwidth}
		\centering
		\caption{With the increase of privacy degree $\epsilon$, CE results of KDDCup99.}
		\label{fig5}
	\end{minipage}
	\hspace{0.02\textwidth}
	\begin{minipage}[t]{0.48\textwidth}
		\centering
		\caption{With the increase of privacy degree $\epsilon$, MSE results of KDDCup99.}
	\end{minipage}
\end{figure}

\begin{figure}[!t]
	\centering
	\begin{minipage}[c]{0.48\textwidth}
		\centering
		\includegraphics[height=4cm,width=7cm,angle=0]{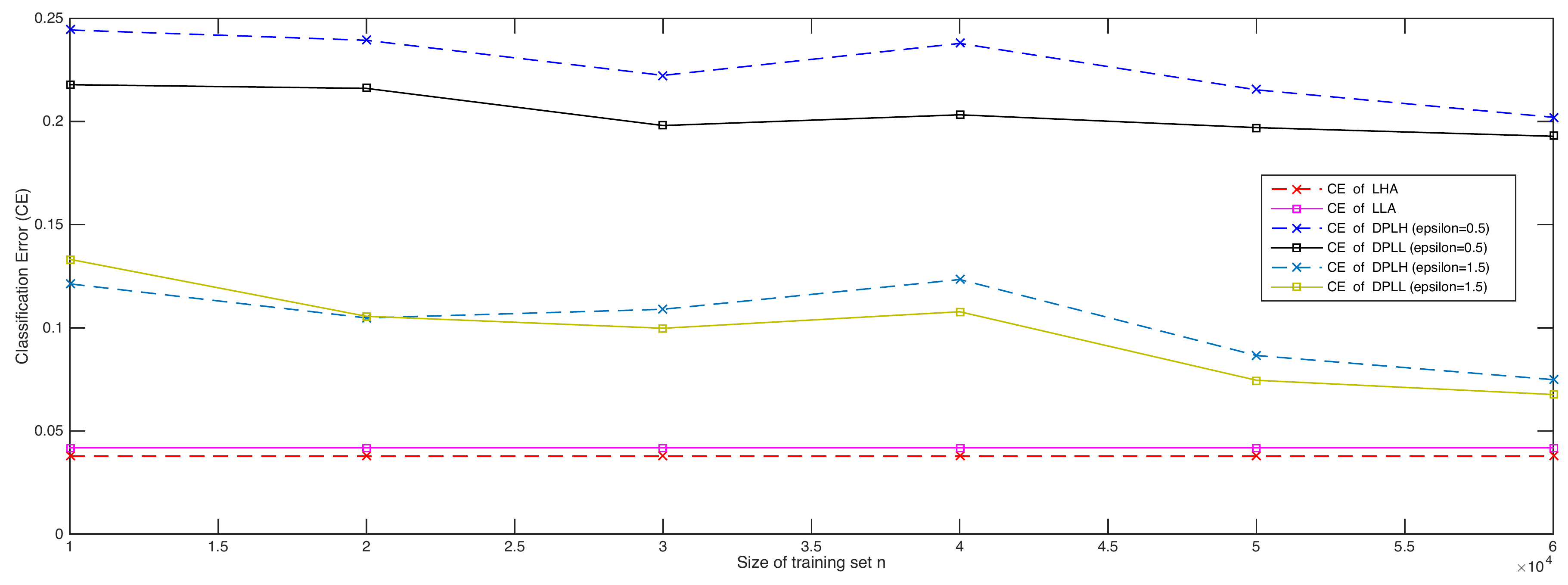}
	\end{minipage}
	\hspace{0.02\textwidth}
	\begin{minipage}[c]{0.48\textwidth}
		\centering
		\includegraphics[height=4cm,width=7cm,angle=0]{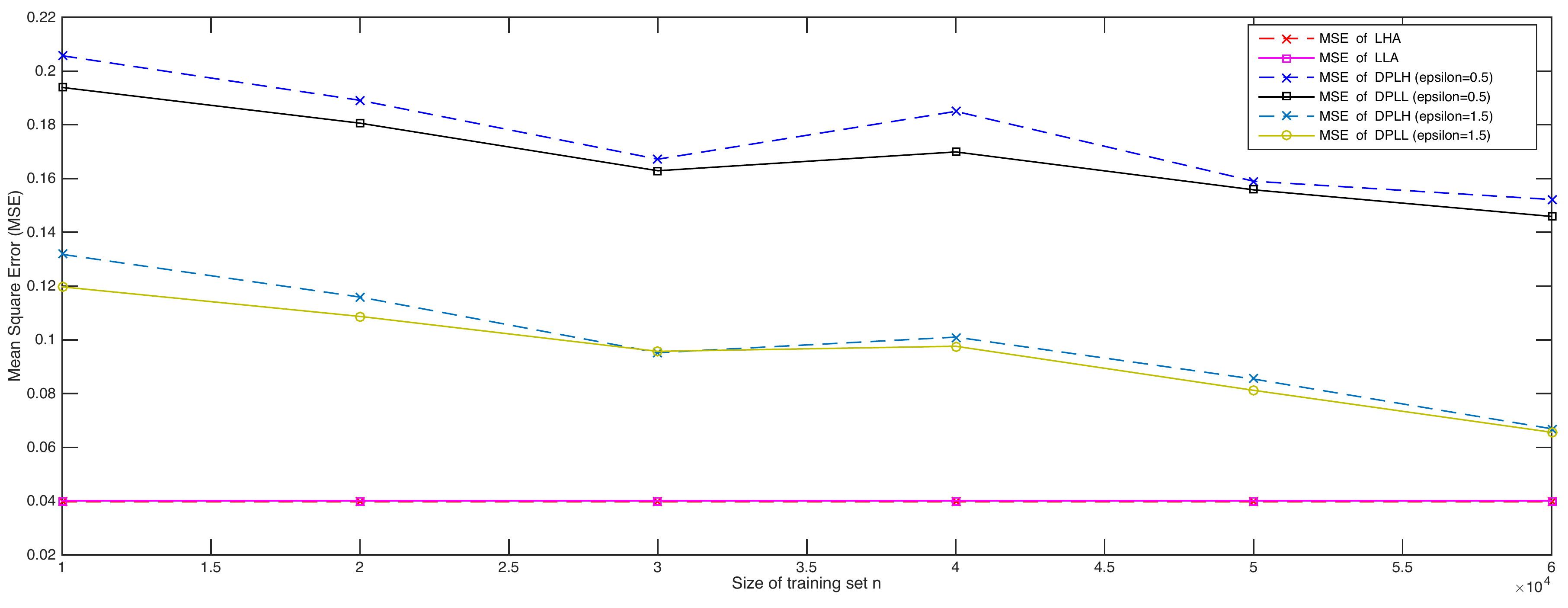}
	\end{minipage}\\[3mm]
	\begin{minipage}[t]{0.48\textwidth}
		\centering
		\caption{With the increase of training data size $n$, CE results of KDDCup99.}
		\label{fig5}
	\end{minipage}
	\hspace{0.02\textwidth}
	\begin{minipage}[t]{0.48\textwidth}
		\centering
		\caption{With the increase of training data size $n$, MSE results of KDDCup99.}
	\end{minipage}
\end{figure}

\section{Conclusions}

In this paper,  
we have proposed a differential privacy version of convex and nonconvex sparse classification approach without stable condition on regularizer. 
Based on ADMM algorithm, 
we transform the solving of sparse problem into a multistep iteration process. 
In each iteration of our algorithm, 
we first deal with an univariate sparse regression which can be solved by the existing sparse algorithm, 
then we solve a data-related $L_2$ regularization problem which is easy to be handled.  
To achieve privacy protecting, we perturb the data-related step in each iteration.
And by the property of the post-processing holding of differential privacy, 
the proposed approach satisfies the $\epsilon-$differential privacy even when the original problem is unstable. 
The theoretical privacy bound of the classification algorithm is $\epsilon\ =\ K\cdot (\frac{2\gamma c_1+2.8c_2}{cn})$. 
The privacy bound of our algorithm is controlled by the algorithm iteration number $K$, the privacy parameter $\gamma$, the parameter of loss function $c_1$ and $c_2$, ADMM pre-selected parameter $c$, and the data size $n$. 
With the increase of privacy parameter $\gamma$, 
the privacy bound $\epsilon$ increases, 
which means that the algorithm leaks more privacy. 
Meanwhile, we perturb the objective functions with a smaller scale. 
Thus, the utility of algorithm increases.  
In addition, as the size of training data $n$ increases, the privacy bound $\epsilon$ decreases, i.e. the possibility of individual privacy leakage decreases. 
It shows that the large traning data make it difficult to distinguish individuals. 
At last, we apply our framework to logistic regression with $L_1$ regularizer and logistic regression with $L_{1/2}$ regularizer. 
Numerical studies  demonstrate that our method is both effective and  efficient which performs well in sensitive data analysis.
The differential privacy sparse classification framework proposed in this work can be easily generalized to other regularization methods, such as SCAD, MCP and so on. 
Furthermore, ADMM algorithm is an effective method to solve distributed problems. Our framework can also be generalized to distributed sparse optimization problems.  
All these problems are under our current research.



{}	


\end{document}